\documentclass{ifacconf}

\usepackage{graphicx}      
\usepackage{natbib}        

\usepackage{xcolor}
\usepackage{enumitem}
\usepackage{float}
\usepackage{subcaption}
\usepackage{algorithm}
\usepackage{algpseudocode}
\usepackage{booktabs}
\usepackage{lipsum}  
\usepackage{amsmath}

\setlist[enumerate]{leftmargin=*} 

\newtheorem{definition}{Definition}

\begin{document}
\begin{frontmatter}

\title{Local Minima Prediction using Dynamic Bayesian Filtering for UGV Navigation in Unstructured Environments\thanksref{footnoteinfo}} 

\thanks[footnoteinfo]{This research is supported in part by the Automotive Research Center (ARC) in accordance with Cooperative Agreement W56HZV-19-2-0001 U.S. Army DEVCOM Ground Vehicle Systems Center (GVSC) Warren, MI. (OPSEC \#: OPSEC8547), and in part by the headquarters of the Republic of Korea Army (ROKA); (Corresponding author:
Dawn M. Tilbury).}

\author{Seung Hun Lee$^2$, Wonse Jo$^1$, Lionel P. Robert Jr.$^{1,3}$, }
\author{and Dawn M. Tilbury$^{1,2,4}$ }

\address{$^1$Robotics, $^2$Electrical Engineering and Computer Science,}
\address{$^3$School of Information, $^4$Mechanical Engineering,}
\address{University of Michigan, Ann Arbor, MI 48109. \\
\{armyhuni,wonse,lprobert,tilbury\}@umich.edu}

\begin{abstract}                
Path planning is crucial for the navigation of autonomous vehicles, yet these vehicles face challenges in complex and real-world environments. Although a global view may be provided, it is often outdated, necessitating the reliance of Unmanned Ground Vehicles (UGVs) on real-time local information. This reliance on partial information, without considering the global context, can lead to UGVs getting stuck in local minima. 
This paper develops a method to proactively predict local minima using Dynamic Bayesian filtering, based on the detected obstacles in the local view and the global goal. 
This approach aims to enhance the autonomous navigation of self-driving vehicles by allowing them to predict potential pitfalls before they get stuck, and either ask for help from a human, or re-plan an alternate trajectory.
\end{abstract}


\end{frontmatter}

\section{Introduction} \label{sec:intro}\vspace{-5pt}
As autonomous vehicles increasingly operate in off-road environments, navigating complex terrains presents significant challenges. A key aspect of this navigation is understanding and addressing the issue of local minima. Local minima can mislead optimization processes by appearing optimal while not leading to globally efficient paths. The local minima problem is particularly prevalent in unstructured environments, where traditional algorithms often struggle to adapt to varying obstacles, elevation changes, and dynamic conditions \citep{meng2024ugv,gong2009robust}. Local path-planning techniques must respond quickly to unforeseen obstacles within their sensory range \citep{sanchez2021path}. One such technique, Artificial Potential Fields (APF), generates obstacle-free paths by assigning positive \textit{``charges''} to distant goals represented as attractive forces ($\vec{F}_{\text{att}}$), and negative charges to nearby obstacles represented as repulsive forces ($\vec{F}_{\text{rep}}$). APF are an attractive solution for local path planning, as they represent a computationally efficient solution for many navigation scenarios, they are intuitive for remote supervisors of the UGVs to understand, and they work really well most of the time \citep{szczepanski2022energy,lazarowska2019discrete,9146273}. 

However, unmanned ground vehicles (UGVs) using APF encounter three significant challenges: becoming trapped in local minima, difficulty navigating between obstacles, and undesirable oscillations \citep{szczepanski2021efficient,orozco2019mobile}. Examples of environments where local minimum problems can occur in the real world include irregular terrain, dense obstacle fields, or complex topographical features (e.g., valleys, cliffs, or narrow passages) \citep{gong2009robust}. Without an effective strategy to avoid or bypass local minima, the UGV may become stuck, not knowing where to move, and fail to reach its goal.

In order to effectively predict local minima, some studies have used well-defined environments, such as fully mapped or convex settings \citep{loizou2011closed, paternain2017navigation,farivarnejad2020design}, or pre-processed environments by transforming into simplified and convex shapes \citep{sawant2023hybrid}, which might result in distortion of the environment.
Other research has focused on adapting to non-convex environments with specific shapes, such as triangular or U-shaped, which can be challenging to apply in a general context \citep{szczepanski2021efficient,szczepanski2022energy}.
Nonetheless, these approaches can be difficult to generalize to the more complex and varied conditions found in real-world settings.




This paper proposes a novel approach to predict local minima in advance using Dynamic Bayesian filtering. Our aim is to leverage the significant computational and logistical advantages of APF as a local-path planning method, while addressing the local minimum challenge through advance prediction.  
We make the following two assumptions: 
A1) the UGV relies on a sensor that provides the distance to local obstacles. This limited perception framework lays the groundwork for our study, aiming to predict the occurrence of local minima in advance within an environment that is only partially known, which is used to mimic real-world off-road environments \citep{nowakowski2023usability}. 
A2) the UGV navigates towards the global goal using its local view and a fixed step size within an APF field based on its current position and the goal location. There exists an attractive force (\(\vec{F}_{\text{att}}\)) towards the goal and repulsive forces (\(\vec{F}_{\text{rep}}\)) from locally observed obstacles. For prediction, we have defined a state transition model that uses raw sensor data to indicate obstacle positions or free space within the sensor range. 
We also consider the uncertainty of the information in its current state to refine the prediction. We demonstrate the superiority of our methodology by comparing its prediction results with existing methods in two common situations where local minima arise.



The main contributions of this paper are as follows:
    \begin{itemize}
       \item We propose a novel method that estimates the probability of the UGV becoming trapped in local minima in unstructured environments using local sensor data; 
       \item We demonstrate that the performance of the proposed method is better than two  existing methods;
    \end{itemize}


The paper is organized as follows. A brief background of previous work done in APF local path planning is discussed in Section \ref{sec:background}. Our proposed method for local minima prediction is presented in Section \ref{sec:method}. The simulation setup and results are detailed in Section \ref{sec:simulation}. Finally, the discussion and conclusion are provided in Sections \ref{sec:discussion} and \ref{sec:conclusion}, respectively, where we briefly discuss how future work can use these predictions of local minima to improve UGV navigation in unstructured and poorly-known environments.

\section{Backgrounds and Related Works} \label{sec:background}\vspace{-5pt}

\subsection{Artificial Potential Field (APF)}
\vspace{-10pt}
The Artificial Potential Field (APF) algorithm operates by considering the sum of virtual forces generated by external elements, namely the global goal position and local obstacles, to determine the motion of a UGV \citep{khatib1986real}. 
The algorithm computes an \(\vec{F}_{\text{att}}\) from the goal position, as well as \(\vec{F}_{\text{rep}}\) from nearby obstacles.

\begin{equation} \label{eq:forces}
    \begin{array}{ll}
    \vec{F}_{\text{att}}(t) &= \xi (X(t) - X_g) \\
    \vec{F}_{\text{rep}}(t) &=
    \sum_{i=1}^{S(t)} \eta \frac{1}{||d_i(t)||^2} \left( \frac{1}{||d_i(t)||} - \frac{1}{\rho_0} \right) \frac{d_i(t)}{||d_i(t)||}
    \end{array}
\end{equation}

\noindent where \(X(t)\) is the UGV's current position at time \(t\), \(X_g\) is the goal position, \(\rho_0\) is the sensor range, \(S(t)\) is the set of obstacles in sensor range at time \(t\), \(d_i(t)\) is the vector from UGV to the \(i\)th obstacle in \(S(t)\), \(\xi\) is the coefficient of $\vec{F}_{\text{att}}(t)$, and $\eta$ is the coefficient of $\vec{F}_{\text{rep}}(t)$.

These virtual forces enable the UGV to reach the goal position by steering it away from obstacles, thereby preventing collisions. The magnitude and direction of these forces depend on the locations of both the goal position and the obstacles. By combining these forces, the UGV is guided away from obstacles while maintaining a safe distance. However, the APF algorithm faces a local minima challenge that occurs when the forces $\vec{F}_{\text{att}}(t)$ and $\vec{F}_{\text{rep}}(t)$ sum to zero.

\subsection{Discrete Time Control}
\vspace{-10pt}
The APF method simplifies the control problem by translating complex environmental interactions into straightforward force vectors. Therefore, the state of the UGV at time $t$ is \(X(t) = [x_t, y_t]\). 
At each time $t$, the UGV moves a distance $\Delta$ in the direction of 
\(\vec{F}_{\text{tot}}(t) = \vec{F}_{\text{att}}(t) - \vec{F}_{\text{rep}}(t) \), as defined in Eq.~\ref{eq:forces}. Let $\theta$ be the direction of $\vec{F_{\text{tot}}}$. Then, the next state of the UGV at time $t+1$ can be described as follows:
\begin{equation}\label{eq:next_state_ugv}
    \begin{array}{ll}
    x_{t+1} &= x_{t} + \Delta \cos\theta \\
    y_{t+1} &= y_{t} + \Delta \sin\theta 
    \end{array}
\end{equation}


\begin{figure}[t]
    \centering
    \includegraphics[width=0.70\linewidth]{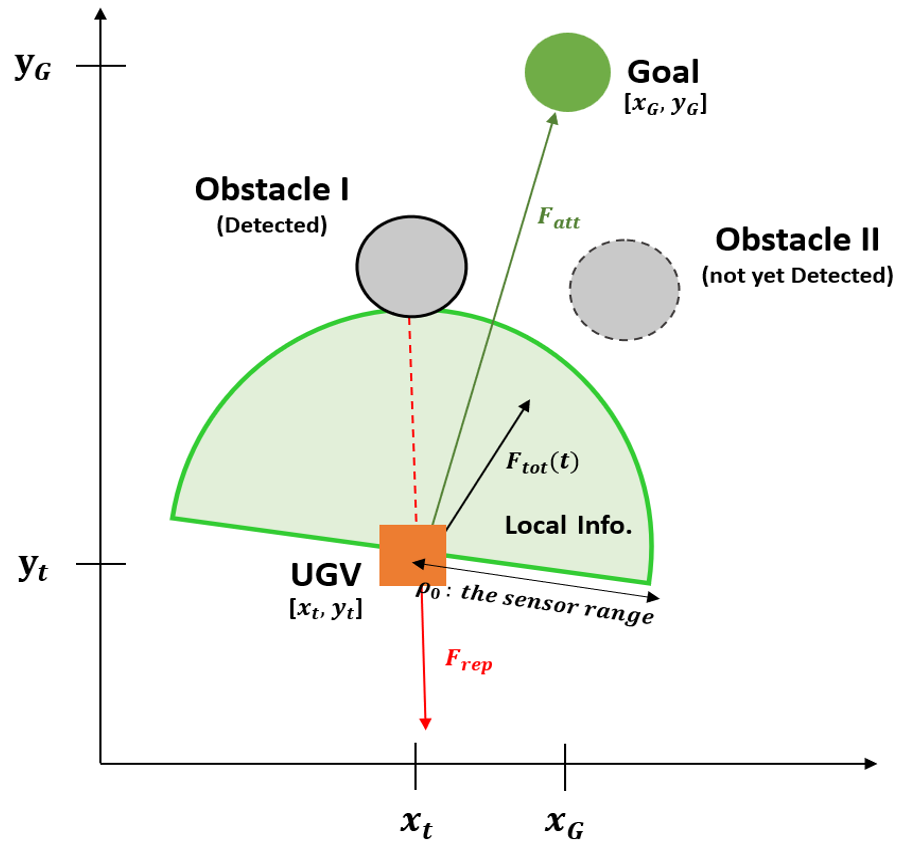}
    \vspace{-10pt}
    \caption{A bird's eye view for UGV navigation using Artificial Potential Field (APF).}
    \label{fig: UGV Navigation in APF}
\end{figure}

\subsection{Local Minima Prediction Methods}\label{sec:related_works}
\vspace{-10pt}
Various methods have been reported to predict local minima in advance. Much of the previous research requires a known or restricted environment that may not be realistic.
Some research has implemented harmonic potentials based on the principle that potential flows in incompressible fluids adhere to Laplace's equation \citep{connolly1997harmonic, masoud2009harmonic, loizou2011closed}. 
Other research has focused on predicting local minima in fully known or convex environments. Several groups created a convex potential field environment and defined the navigation function to avoid local minima \citep{paternain2017navigation, vasilopoulos2018reactive, sawant2023hybrid}. 
Arslan \textit{et al.} found that there is one unique attractor at the goal location and one local minimum point associated with each obstacle among the stationary points in known convex sphere worlds \citep{arslan2019sensor}. 

In contrast, recent research has focused on non-convex environments, which resemble complex and realistic settings, and predicted the local minima from the U-shaped obstacle by using the shape of a LiDAR sensor's range or from the triangular shape by using the horizon of a LiDAR sensor \citep{szczepanski2021efficient,szczepanski2022energy, rodriguez2024reactive,jin2021hierarchical}. 
However, this method is difficult to generalize for all shapes and real-world environments. This method also fails to account for unknown areas when updating the predictions, since the prediction will become more accurate as more information becomes available. The APF algorithm still faces the local minima challenge that occurs when the forces $\vec{F}_{\text{att}}$ and $\vec{F}_{\text{rep}}$ sum to zero.

\section{Dynamic Bayesian Filtering-based Local Minima Prediction Algorithm} \label{sec:method}\vspace{-5pt}
In this section, we propose a novel approach for predicting a local minimum using Dynamic Bayesian filtering in APF. In Section \ref{sec:local_minimum}, we introduce the conditions where the local minimum occurs in APF. In addition, we explain how the area information can be described based on the sensing range of the UGV. Finally, we demonstrate how the belief of a local minimum can be recursively updated using the area information.

\begin{definition}[Local Minimum (\(X_{lm}\))]
A local minimum (\(X_{lm}\)) is a position where the attractive force (\(\vec{F}_{\text{att}}(t)\)) and the repulsive force (\(\vec{F}_{\text{rep}}(t)\)) are balanced; i.e., the total force vector \(\vec{F}_{\text{tot}}(t)\) is zero.
\end{definition}

\subsection{Local Minimum Prediction} \label{sec:local_minimum}
\vspace{-10pt}
A necessary condition for a local minimum to exist is that \( \vec{F}_{\text{att}}(t) \) and \( \vec{F}_{\text{rep}}(t)\) are parallel. If the UGV moves along this vector field, the repulsive force from the obstacle(s) will increase (and the attractive force from the goal will decrease), and the UGV could 
encounter a minimum point where the forces are equal and opposite (but non-zero). On the other hand, as the UGV moves along the vector field,
it could encounter another obstacle that would add to the repulsive force (See Obstacle II in Fig.~\ref{fig: UGV Navigation in APF}) and potentially
break the parallelism. In this case, the local minima may not exist as originally predicted. Therefore, the potential presence of obstacles in unknown areas is important to consider when predicting a local minimum. 

Here, we assume the UGV is moving toward its goal with a sensor oriented in the UGV's heading direction and navigating an environment with unknown obstacles on its path. At each time step $t$, it can sense an area in front of it, identifying obstacles within a defined radius. Using this \textit{Local.Info} and the location/direction of obstacles, it can compute \( \vec{F}_{\text{rep}}(t)\). If \( \vec{F}_{\text{att}}(t) \) and \( \vec{F}_{\text{rep}}(t) \) are parallel, then Algorithm 1 is invoked to predict a local minimum.

\begin{definition}[Sensing Area (SA(X))]
    An area equivalent to Local.Info(t) when the UGV is positioned at X(\(t\)) (see Fig.~\ref{fig: TRA, AOI, and UA Figures}).
\end{definition}

\begin{definition}[Area Definitions]
In the case when the attractive and repulsive forces are parallel, three distinct areas are defined in relation to the UGV's navigation and obstacle detection capabilities:
    \begin{itemize}
        \item \textbf{Recognized Area (RA($X_t$))}: An area in AOI that is recognized by the UGV up to time \(t\) (i.e. $\bigcup_{X=X_{t_0}}^{X_{t}} SA(X)$).\\

        \item \textbf{Area of Interest (AOI)}: An area identified as an entire area that encompasses from $RA(X_{t_0})(=SA(X_{t_0}))$ to $SA(X_{lm})$ (i.e. $\bigcup_{X=X_{t_0}}^{X_{lm}} SA(X)$).\\
    
        \item \textbf{Unknown Area (UA($X_t$))}: An area within AOI that has not been recognized by the UGV at time \(t\).
    \end{itemize}
\end{definition}

\begin{algorithm}
\caption{Local Minima Prediction}
\begin{algorithmic}
\State Assume \( \vec{F}_{\text{att}}(t) \) and \( \vec{F}_{\text{rep}}(t) \) are parallel at time \(t_0\), when the UGV is at \(X_{t_0}\)

\\
\State \textbf{(Initialization Step; see Fig.~\ref{fig: TRA, AOI, and UA Figures})}

\State $RA(X_{t_0}) \gets Local.Info(t_0)$
\State $X_{lm} \gets$ the projected point where $F_{tot} = 0$
\State $SA(X_{lm}) \gets$ UGV could observe from $X_{lm}$
\State $AOI \gets RA(X_{t_0}) \cup SA(X_{lm})$

\State Generate Group A 
\State Define initial belief $bel_{t_0}(X_{lm})$
\\
\State \textbf{(Recursive Step)}
\While{$bel_t(X_{lm}) < \gamma$}
    \State 1. Prediction Step:
    \State Update $\bar{bel}(X_{lm})$ by multiplying state transition 
    \State probability (Definition 4)
    \State 2. Correction Step:
    \State Refine $bel(X_{lm})$ by multiplying observation 
    \State likelihood (Definition 5)
    \State 3. Normalization Step:
    \State Adjust $bel(X_{lm})$ by multiplying normalization 
    \State factor $\nu$
    \If{$bel(X_{lm}) < \gamma$ at time $t$}
        \State $ t \gets t+1 $     
        \State UGV moves one step to $X(t + 1)$ along $\vec{F}_{\text{tot}}$
    \Else
        \State break
    \EndIf
\EndWhile

\State \State Report: \textit{``UGV is likely to get stuck in $X_{lm}$ with $\gamma \times 100 \%$ confidence level.''}
\end{algorithmic}
\label{alg:prediction}
\end{algorithm}

As the UGV traverses along the APF, if no further obstacles are detected, the local minimum will eventually be reached, and the UGV would stop. In this case, the higher-level planning algorithm could attempt an avoidance maneuver. 
On the other hand, if more obstacles are detected, the parallelism of the attractive and repulsive vectors could be broken, and a local minimum will not occur, and the UGV could continue traversing towards the goal along the APF.

To clarify which areas are known or unknown at time \(t\), we categorize the area around the UGV into three distinct zones: AOI, RA($X_t$), and UA($X_t$) (see Definition 3). First, the AOI is identified after \( \vec{F}_{\text{att}}(t) \) and \( \vec{F}_{\text{rep}}(t) \) are parallel at time \(t_0\), indicating that a local minimum is likely to appear within the AOI. Second, the RA($X_t$) is the area within which the UGV can detect obstacles and free spaces for maneuvering up to time \(t\), and it increases as the UGV moves forward. Lastly, the UA($X_t$) is an area within the AOI that the UGV cannot recognize at time \(t\), and it decreases over time after \(t_0\). In the UA($X_t$), undiscovered obstacles may cause the UGV to encounter local minima sooner or allow it to maneuver toward the goal without encountering the initial local minima within a few steps.

\begin{figure}[t]
    \centering
    \includegraphics[width=0.8\linewidth]{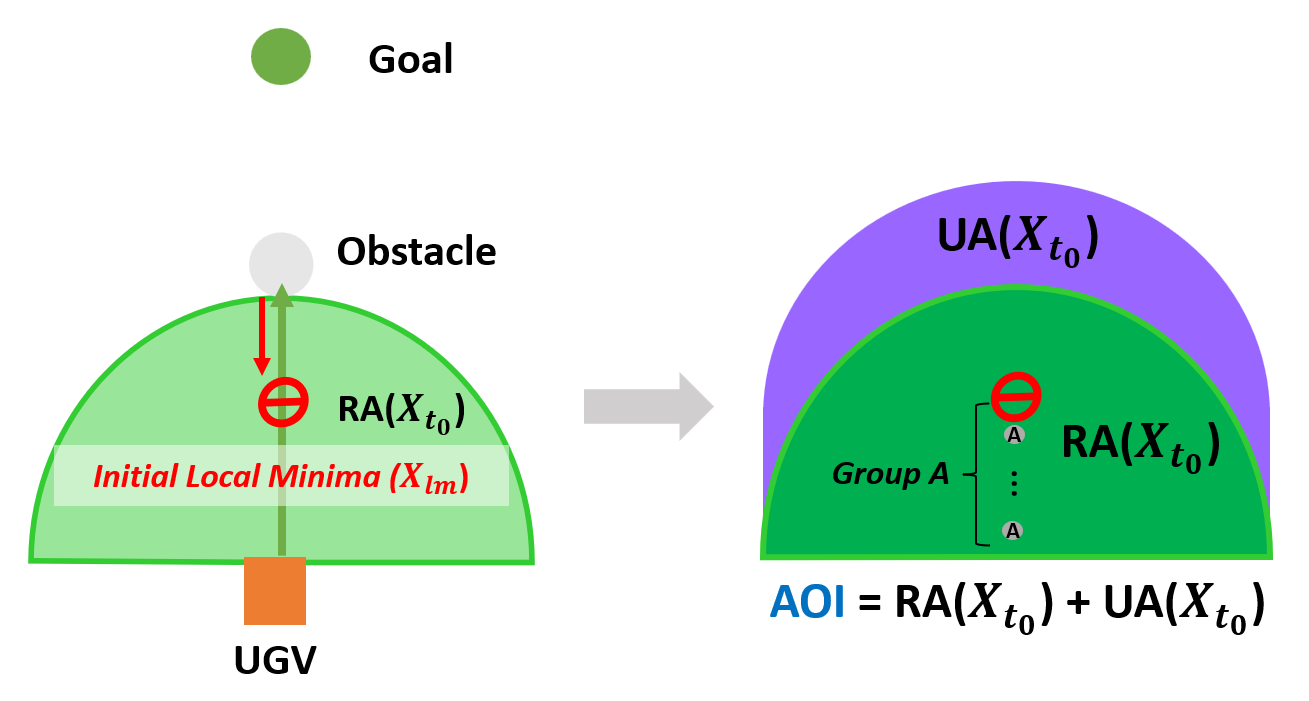}
    \vspace{-10pt}
    \caption{Examples of AOI, RA($X_{t_0}$), and UA($X_{t_0}$) used in the proposed local minima prediction algorithm.}
    \label{fig: TRA, AOI, and UA Figures}
\end{figure}

In Algorithm \ref{alg:prediction}, we can identify the initial \(X_{lm}\), where \( \vec{F}_{\text{total}}(t) \) is predicted to be zero in RA($X_{t_0}$) if no other obstacles are encountered. Moreover, by dividing the line between the UGV and the initial \(X_{lm}\), we can generate potential points, denoted as Group A. The points outside Group A, which lie between the obstacle and the initial local minimum, represent points where the UGV cannot maneuver within the APF field, as \( \vec{F}_{\text{rep}}(t) \) from the obstacles outweighs \( \vec{F}_{\text{att}}(t) \). However, Group A comprises points between the initial local minimum and the UGV, including the local minimum itself, where the UGV may be able to maneuver within the APF field since \( \vec{F}_{\text{att}}(t) \) exceeds \( \vec{F}_{\text{rep}}(t)\). These points are likely to become local minima unless the UGV encounters undetected obstacles that change the direction of \( \vec{F}_{\text{rep}}(t)\) in the UA($X_t$). Therefore, we can define the initial belief (\(bel_{t_0}(X_{lm})\)), which represents the probability that the initial \(X_{lm}\) could be the local minimum among the points in Group A: \(bel_{t_0}(X_{lm}) = 1/|\text{Group A}| \), where $|\text{Group A}|$ is the number of points in Group A. As noted in Sec.~\ref{sec:discussion}, there may be other ways to define this initial belief.

\subsection{Dynamic Bayesian Filtering}
\vspace{-10pt}
In Dynamic Bayesian filtering \citep{sarkka2023bayesian}, the filtering process involves three sequential steps: prediction, correction, and normalization under the Markov Assumption.
The prediction step uses the system's prediction model to forecast the subsequent state by incorporating the state transition probabilities and the prior belief from the previous time step. In the correction step, the prediction is refined with the latest observations \(z_t\) (i.e. the UGV perceives $X_{lm}$ as such in RA($X_{t}$)) by computing the likelihood of the new observation given the predicted state, resulting in a corrected posterior belief. Finally, the normalization step adjusts the posterior belief so that the sum of the probabilities equals one, thus maintaining a valid probability distribution.

\subsubsection{Prediction Step}

\begin{definition}[State Transition Probability]
The state tr-ansition probability, denoted by \(P(X_{lm}^t \mid X_{lm}^{t-1}, z_{1:t-1}, u_{1:t})\), represents the probability of the status (being a local minimum) of \(X_{lm}\) at time \(t\) given its status at time \(t-1\), the sequence of all observations up to time \(t-1\), and the sequence of all inputs up to time \(t\). The input \(u_t\) indicates that the UGV follows the total force \(\vec{F}_{\text{total}}(t)\). Under the Markov Assumption, \(P(X_{lm}^t \mid X_{lm}^{t-1}, z_{1:t-1}, u_{1:t})\) is equivalent to \(P(X_{lm}^t \mid X_{lm}^{t-1}, u_t)\).
\end{definition}

\begin{equation}
    \begin{array}{ll}
    P(X_{lm}^t | X_{lm}^{t-1}, u_{t}) &= \frac{\alpha}{\pi}
    \end{array}
\end{equation}

\begin{figure}[t]
    \centering
    \begin{subfigure}{0.49\linewidth}
        \centering
        \includegraphics[width=1\linewidth]{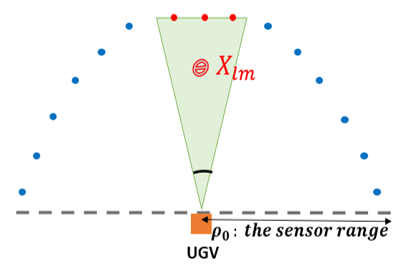}
        \vspace{-15pt}
        \caption{Case 2}
    \end{subfigure}
    \begin{subfigure}{0.49\linewidth}
        \centering
        \includegraphics[width=1\linewidth]{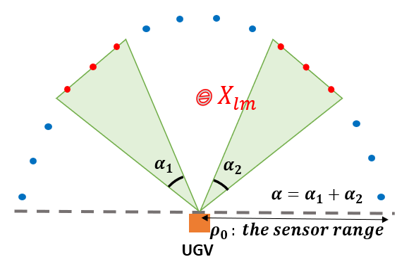}
        \vspace{-15pt}
        \caption{Case 2}
    \end{subfigure}    
    \vspace{-20pt}
    \caption{Occupied points angle in (a) Case 1 - Wall and (b) Case 2 - Hallway.}
    \label{fig: Occupied Points Angle}
\end{figure}

We can predict whether the UGV will become stuck at the initial \(X_{lm}\) based on data points within the sensor's range ($\pi$), which indicates the presence of obstacles or free space. We use the angle of occupied points ($\alpha$), which refers to the ratio of the number of occupied points (represented by red dots in Fig.~\ref{fig: Occupied Points Angle}) to the total number of sensor points. A higher ratio suggests that obstacles are present nearby the UGV, increasing the likelihood that the UGV will become trapped at the initial \(X_{lm}\).

\subsubsection{Correction Step}

\begin{definition}[Observation Likelihood]
Observation likelihood, denoted by \(P(z_{t} | X_{lm}^{t}, z_{1:t-1}, u_{1:t})\), represents the probability of observing the state \(z_t\) given the expected local minima point (\(X_{lm}^{t}\)), the observations (\(z_{1:t-1}\)) and the inputs (\(u_{1:t}\)). 
This likelihood can be interpreted as the information that the UGV possesses in AOI at time \(t\). Under the Markov Assumption, \(P(z_{t} | X_{lm}^{t}, z_{1:t-1}, u_{1:t})\) is equivalent to \(P(z_{t} | X_{lm}^{t})\).
\end{definition}

\begin{equation}
    \begin{array}{ll}
    P(z_t \mid X_{lm}^{t} = \text{Local Min}) = \frac{\text{RA($X_t$)}}{\text{AOI}}
    \end{array}
\end{equation}

During the correction step, the prediction $\bar{bel}_t(X_{lm})$ is refined using the recent observation $z_t$. This refinement involves leveraging the area information that the UGV perceives in AOI in order to identify a local minimum, i.e. the ratio of RA($X_{t}$) to AOI. This process leads \(\bar{bel}_t(X_{lm})\) to an updated posterior belief by multiplying with observation likelihood which represents the probability of the new observation based on the predicted state \(X_{lm}^{t}\). This adjustment ensures that the belief becomes more accurate by incorporating newly acquired information.

\subsubsection{Normalization} 
In the normalization step, the posterior belief is adjusted to ensure that the total probabilities sum to one, thereby preserving a valid probability distribution. Here, \(\nu\) represents the normalization factor.


After the normalization process, the normalized belief \( \text{bel}_t(X_{lm} = \text{Local Min}) \) is obtained. We use a predefined threshold \( \gamma \) to determine when the UGV should halt its navigation to avoid getting stuck in \( X_{lm} \). If \( \text{bel}_t(X_{lm} = \text{Local Min}) \) exceeds \( \gamma \), it implies that the UGV has a confidence level corresponding to \( \gamma \times 100\% \) that \( X_{lm} \) is a local minimum and is likely to get stuck within a short number of steps. The estimated number of steps before the UGV reaches \( X_{lm} \) can be calculated by dividing the distance.

\section{Simulation Test and Results} \label{sec:simulation}\vspace{-5pt}

\subsection{Simulation Setup} 
\begin{figure}[t]
    \centering
    \begin{subfigure}{0.4\linewidth}
        \centering
        \includegraphics[width=1\linewidth]{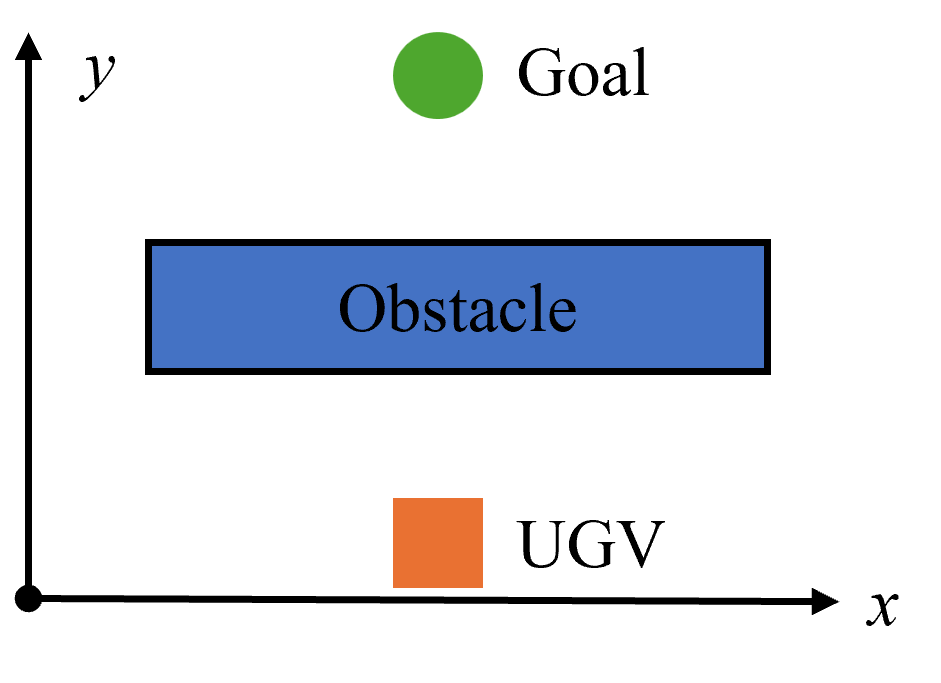}
        \vspace{-20pt}
        \caption{Case 1}
    \end{subfigure}
    \begin{subfigure}{0.4\linewidth}
        \centering
        \includegraphics[width=1\linewidth]{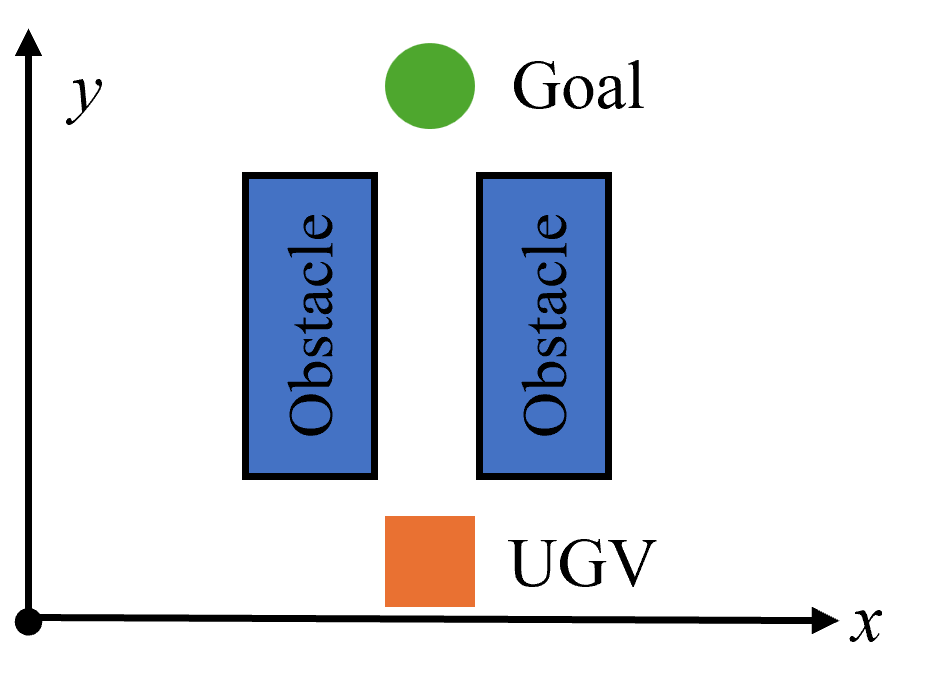}
        \vspace{-20pt}
        \caption{Case 2}
    \end{subfigure}
    \vspace{-10pt}
    \caption{Two occupancy simulation environments: (a) Case 1 - Wall and (b) Case 2 - Hallway.}
    \label{fig: Two Occupancy Maps for Simulations}
\end{figure}

\vspace{-10pt}
Our novel approach was implemented in MATLAB \citep{matlab_robot} to evaluate its effectiveness. A simulation was created to implement a UGV (represented by an orange box) navigating through an APF field towards a goal location (represented by a green circle). 
%
%
Simulations were conducted in two common scenarios of local minimum occurrence for UGVs: 
Case 1) the UGV is blocked by a long, wall-shaped obstacle, and Case 2) the UGV is unable to find a path due to the effects of repulsive forces such as through a hallway (See Fig.~\ref{fig: Two Occupancy Maps for Simulations}). We conducted our simulation using a 3D occupancy map. For sensor representation, we used the ray-object intersection, which provides binary results indicating whether the space is occupied or free in the occupancy map, and the distance to the obstacles if they exist.

\subsection{Comparison Methods}
\begin{figure}[t]
    \centering
    \begin{subfigure}{0.4\linewidth}
        \centering
        \includegraphics[width=1\linewidth]{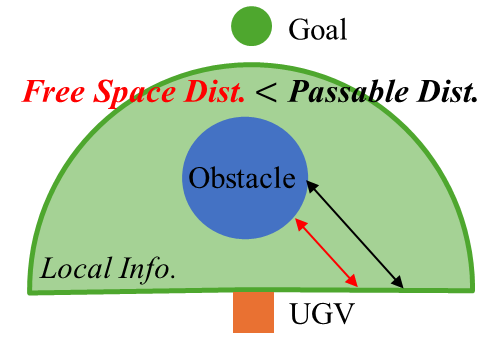}
        \vspace{-15pt}
        \caption{Method 1}
    \end{subfigure}
    \begin{subfigure}{0.4\linewidth}
        \centering
        \includegraphics[width=0.9\linewidth]{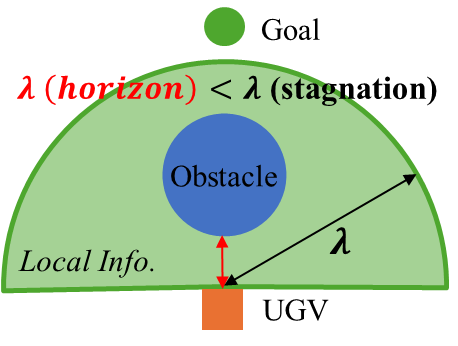}
        \vspace{-3pt}
        \caption{Method 2}
    \end{subfigure}   
    \vspace{-10pt}
    \caption{Two comparison methods: (a) Method I and (b) Method II.}
    \label{fig: State Transition Probability}
\end{figure}

\vspace{-10pt}
We compared our methodology with two recent research studies focused on non-convex environments within partially known settings (see Fig.~\ref{fig: State Transition Probability}). 
Method I predicts the local minimum by examining the distance between adjacent points in the free space in Lidar data. If, within the analyzed range, there is no passage for the robot to navigate through, then the algorithm identifies a local minimum \citep{szczepanski2021efficient}. 
Method II predicts the local minimum by checking the horizon distance along the extended line of $\vec{F}_{\text{tot}}$ from the robot's position to the nearest obstacle. If the horizon distance $\lambda_{\text{horizon}}$ is less than half of the Lidar's range $\lambda_{\text{stagnation}}$, the algorithm predicts a local minimum \citep{szczepanski2022energy}. We evaluated whether their prediction algorithms were successful in two common local minimum scenarios (see Fig.~\ref{fig: Two Occupancy Maps for Simulations}) and assessed how far in advance they could predict the local minimum.  

\renewcommand{\arraystretch}{1} 
\begin{table}[h]
    \centering
    \caption{Time steps to local minima detection}
    \label{tab:comparison}
    \vspace{-5pt}
    \begin{tabular}{l|c|c}
    \toprule
    \textbf{Methods} & \textbf{Case 1: Wall} & \textbf{Case 2: Hallway} \\
    \hline     
    Method I & 108 & 63 \\
    Method II & 108 & 63 \\
    \textbf{Novel Method} & \textbf{100} & \textbf{54} \\
    \toprule

    \end{tabular}

\end{table}
\renewcommand{\arraystretch}{1}




\subsection{Results}
\vspace{-10pt}
Both comparison methods failed to stop in advance or to predict the occurrence of a local minimum, see Table \ref{tab:comparison}. The UGV halts its navigation when it encounters the local minimum at the 108th step in Case 1 and at the 63rd step in Case 2, as defined in Definition 1.

For our novel method, simulations were conducted with the threshold \(\gamma\) $=$ 0.85. If \(\text{bel}_t(X_{lm} = \text{Local Min})\) exceeds \(\gamma\), it indicates that the initial \(X_{lm}\) is likely to become a local minimum with a probability of 85\%. Our novel approach can predict the local minimum in advance in both cases, not only identifying the location of the local minimum but also the number of steps remaining before the UGV reaches the predicted local minimum. 

\begin{figure}[t]
    \centering
    \begin{subfigure}{0.49\linewidth}
        \centering
        \includegraphics[width=1\linewidth]{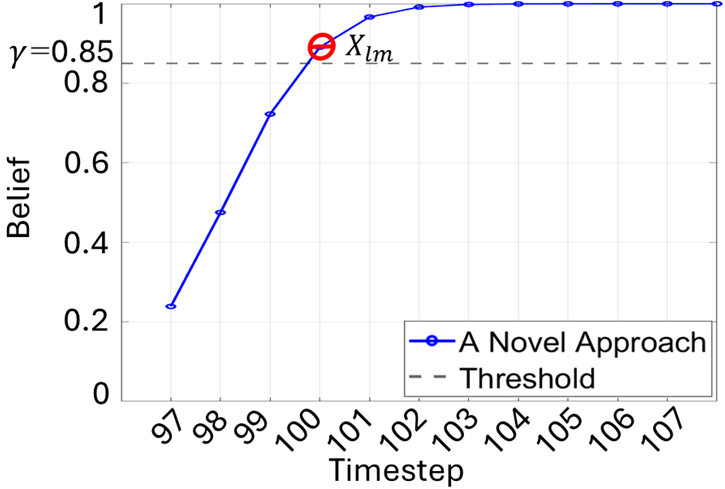}
        \vspace{-15pt}
        \caption{Case 1}
        \label{fig:case1}
    \end{subfigure}   
    \begin{subfigure}{0.49\linewidth}
        \centering
        \includegraphics[width=1\linewidth]{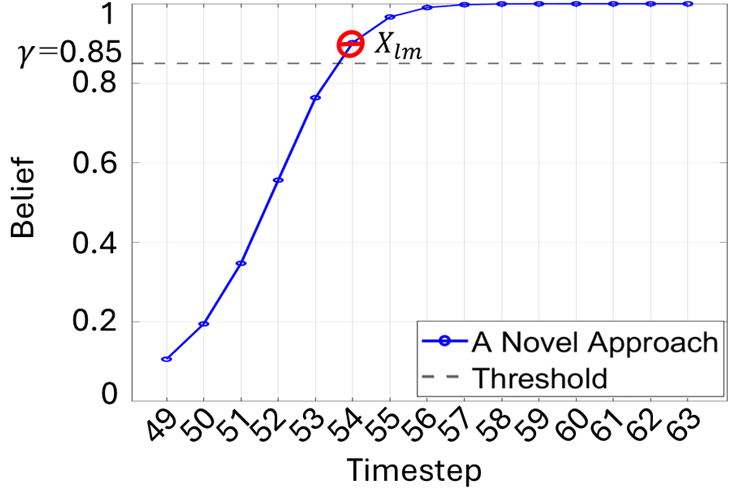}
        \vspace{-15pt}
        \caption{Case 2}
        \label{fig:case2}
    \end{subfigure}   
    \vspace{-20pt}
    \caption{Belief in local minimum in (a) Case 1 - Wall and (b) Case 2 - Hallway.}  
    \label{fig:both_cases}
\end{figure}

In Case 1, our approach can halt the UGV at the 100th step with an 89\% confidence level. In Case 2, the UGV halts its navigation at the 54th step with a 90\% confidence level; see Fig.~\ref{fig:both_cases}. This early prediction of the local minimum with high confidence enables other options to be considered in a timely fashion. There is no need to  just wait until the UGV stops to consider alternate plans.

\section{Discussion} \label{sec:discussion} \vspace{-5pt}

In this paper, we proposed a local minima prediction algorithm in local path planning for UGVs. The results demonstrated that our methodology successfully calculates the probability of local minima occurrence, with its accuracy progressively improving over time. This approach offered significant advantages over existing methods in terms of handling uncertainties in unknown environments. 

However, in this study, we found several limitations in our methods of predicting local minimum from limited environmental information. 
The first limitation is that although our simulation represents the environment navigated by the UGV in off-road settings using 3D occupancy map rather 2D one, it has not been validated with a UGV in real-world environments due to limited lab facilities and safety concerns. However, we plan to address this limitation by testing our method in a new game-engine-based realistic simulation environment (e.g., Unreal) as part of our future work.
Second, we validated our local minima prediction algorithm with the simplified kinematic UGV model in the simulation environment, which is the differential wheel drive model of a non-holonomic UGV \citep{siegwart2011introduction}. Given the real-world environment, we should utilize the advanced kinematic UGV models to improve the accuracy of the prediction.
Third, the initial belief of a local minimum, which we assumed to follow a uniform distribution, could instead adhere to other distributions with a higher initial probability that reflects the complexity of the environment. 
Fourth, for time series analysis to be applicable in real-world scenarios, we assumed that observations at each time step are independent, in accordance with the Markov Assumption. However, in real settings, the observation at time \(t\) can be correlated with both previous and future time steps.
Lastly, we did not consider how to overcome local minima to arrive at the goal since the objective of this study is to predict local minima in advance, so we assumed that the UGV could escape the local minimum after getting stuck by making random movements. 

Future research could consider employing this novel methodology to prevent the UGV from being trapped in local minima by requesting the intervention of a human operator. The UGV can significantly benefit from human intervention since a human operator possesses adaptive and holistic reasoning abilities, influenced by their experiences and knowledge across multiple domains.
Leveraging their knowledge and experience, the human operator could suggest waypoints that enable the UGV to navigate through areas previously perceived by the UGV, or suggest promising directions for the UGV to explore.

\section{Conclusion} \label{sec:conclusion} \vspace{-5pt}

This paper proposes a novel methodology for predicting local minima based on Dynamic Bayesian Filtering. The results demonstrate that, compared to previous studies, our methodology enables the UGV to predict local minima in advance. Furthermore, the state transition probability from an initial local minimum point enables the UGV to predict subsequent local minima in advance. This not only enhances the post-belief of a local minimum but also allows the UGV react earlier with greater confidence. Future work will include higher-level planning methods that can avoid local minima by asking human supervisors for suggestions.

\begin{ack}\vspace{-5pt}
The authors would like to thank Gen Sasaki from MathWorks Inc., and Kayla Riegner and Jonathon Smereka from U.S. Army DEVCOM Ground Vehicle Systems Center (GVSC) for their valuable insights and feedback throughout the development of this work.
\end{ack}

\bibliography{ifacconf}             
                                                   







\end{document}